%% file: main.tex
\documentclass[11pt]{article}
\usepackage[preprint]{acl}
\usepackage[T1]{fontenc}
\usepackage[utf8]{inputenc}
\usepackage{times}
\usepackage{latexsym}
\usepackage{microtype}
\usepackage{graphicx}
\usepackage{booktabs}
\usepackage{amsmath}
\usepackage{amssymb}
\usepackage{multirow}
\usepackage{xcolor}
\usepackage{CJKutf8}

\newcommand{\zh}[1]{#1}
\input{tables/result_macros}

\title{Mandarin Wh-Indefinites Follow Nonveridicality in Multilingual LLMs, Unlike English \emph{any}}

\author{Guofang Tang \\
  \texttt{tinggofun@gmail.com}}

\begin{document}
\begin{CJK}{UTF8}{gbsn}
\maketitle

\input{sections/00_abstract}
\input{sections/01_intro}
\input{sections/02_setup}
\input{sections/03_results}
\input{sections/04_ratings}
\input{sections/05_mech}
\input{sections/06_related}
\input{sections/07_discussion}

\bibliography{refs}

\appendix
\input{sections/09_appendix}

\end{CJK}
\end{document}

%% file: tables/result_macros.tex
\newcommand{\AgreeEnMax}{0.85}
\newcommand{\AgreeEnMin}{0.72}
\newcommand{\AgreeZhMax}{0.53}
\newcommand{\AgreeZhMin}{0.44}

\newcommand{\BigLenEA}{0.31}
\newcommand{\BigLenZAmax}{0.05}
\newcommand{\BigLzhZAmax}{1.92}
\newcommand{\BigLzhZAmin}{0.85}

\newcommand{\BigNEA}{1}
\newcommand{\BigNZA}{3}

\newcommand{\ChkFZhMin}{0.90}

\newcommand{\ChkOZhMax}{0.89}
\newcommand{\ChkOZhMin}{0.57}
\newcommand{\FcRhoMax}{0.92}
\newcommand{\FcRhoMin}{0.58}
\newcommand{\LenEAmax}{0.72}
\newcommand{\LenEAmin}{0.10}
\newcommand{\LenZAmax}{0.15}
\newcommand{\LenZAmin}{$-$0.26}
\newcommand{\LzhZAmax}{2.33}
\newcommand{\LzhZAmin}{0.42}
\newcommand{\MeanLFirstEn}{0.54}
\newcommand{\MeanLFirstZh}{0.35}
\newcommand{\MeanLOnlyEn}{0.35}
\newcommand{\MeanLOnlyZh}{0.30}
\newcommand{\MeanLRegretEn}{0.25}
\newcommand{\MeanLRegretZh}{0.13}
\newcommand{\NAddEnen}{10}
\newcommand{\NAddEnzh}{3}
\newcommand{\NAddZhen}{9}
\newcommand{\NAddZhzh}{1}
\newcommand{\NAdvAll}{66}
\newcommand{\NAdvPos}{66}
\newcommand{\NCatEFNeither}{7}

\newcommand{\NCatEFShared}{3}
\newcommand{\NCatEONeither}{6}

\newcommand{\NCatFAssess}{11}
\newcommand{\NCatFNeither}{7}

\newcommand{\NCatFPartial}{1}
\newcommand{\NCatFSeparate}{2}
\newcommand{\NCatFShared}{1}
\newcommand{\NCatOAssess}{7}

\newcommand{\NCatONotassessable}{4}

\newcommand{\NEnenAbove}{11}
\newcommand{\NEnenEAAbove}{8}
\newcommand{\NEnzhAbove}{4}
\newcommand{\NEnzhZAAbove}{4}
\newcommand{\NFcEvidPos}{7}
\newcommand{\NFcFirstNeg}{7}
\newcommand{\NFcGate}{8}
\newcommand{\NFcNecPos}{7}
\newcommand{\NFcOnlyNeg}{8}
\newcommand{\NFcPossPos}{7}
\newcommand{\NFcRegretNeg}{6}
\newcommand{\NFirstDeltaNeg}{7}

\newcommand{\NFirstLenPos}{11}
\newcommand{\NMech}{11}
\newcommand{\NModels}{12}
\newcommand{\NMpass}{11}

\newcommand{\NOnlyDeltaNeg}{5}

\newcommand{\NOnlyLenPos}{10}
\newcommand{\NRegretDeltaNeg}{5}

\newcommand{\NRegretLenPos}{11}
\newcommand{\NRobAll}{165}
\newcommand{\NRobPos}{165}

\newcommand{\NZhzhBoth}{7}

\newcommand{\NenSide}{3}
\newcommand{\NzhSide}{11}
\newcommand{\QthreeNegDiD}{0.47}
\newcommand{\QthreeNegFrac}{60}
\newcommand{\QthreeNegHi}{1.07}
\newcommand{\QthreeNegLo}{$-$0.14}
\newcommand{\RateBaseEn}{1.4}

\newcommand{\RateDidRhoEn}{0.67}
\newcommand{\RateDidRhoZh}{0.80}
\newcommand{\RateFirstEn}{4.0}
\newcommand{\RateFirstZh}{1.9}
\newcommand{\RateOnlyEn}{4.1}
\newcommand{\RateOnlyZh}{2.5}
\newcommand{\RateRegretEn}{3.9}
\newcommand{\RateRegretZh}{2.8}
\newcommand{\RateRhoEn}{0.72}
\newcommand{\RateRhoZh}{0.80}
\newcommand{\RenenMax}{0.76}
\newcommand{\RenenMin}{0.21}
\newcommand{\SmallLenEAmax}{0.45}
\newcommand{\SmallLenEAmin}{0.28}
\newcommand{\VvRhoMin}{0.979}

%% file: sections/00_abstract.tex
\begin{abstract}
Multilingual LLMs learn mostly from English. Do they carry English polarity licensing over to other languages?
English \emph{any} is licensed by downward-entailing (DE) contexts, Mandarin wh-indefinites such as \zh{什么}
`something' by nonveridical ones, and the two classes come apart: \emph{perhaps} licenses \zh{什么} but not
\emph{any}, while \emph{only} licenses \emph{any} but entails existence. We test this dissociation in a
pre-registered study of twelve open LLMs (0.5B--9B, four families), using 48 matched bilingual frames in 21 contexts
and a log-probability difference-in-differences normalised by negation. All \NMpass{} models that pass the
measurement gate license Mandarin wh-indefinites under possibility, necessity and evidential adverbs, at
\LzhZAmin--\LzhZAmax{} times the strength of negation, while English \emph{any} gets nothing in the same contexts.
An exploratory extension to Qwen2.5-32B shows the same.
The English side is weaker: \emph{any} under \emph{only}, \emph{first} and \emph{regret} is licensed but at a
fraction of negation, so only \NenSide{} models show the full double dissociation in sentence probabilities. When
prompted to choose between sentences, most gated models show it. In exploratory directional ablations, a direction
that carries English licensing, held-out classes included, does not reach Mandarin licensing in most models. The
models do not simply apply the English rule to Mandarin.
\end{abstract}

%% file: sections/01_intro.tex
\section{Introduction}

Multilingual LLMs learn most of what they know from English, and their internal representations are partly
English-anchored \citep{wendler2024llamas}. For a phenomenon that exists in many languages but works differently in
each, this raises a concrete question. Does the model learn each language's own rule, or does it learn the English
rule and apply it everywhere?

Polarity-sensitive indefinites are a sharp test case. English \emph{any} is licensed by downward-entailing (DE)
operators: \emph{Tom didn't buy anything} is fine, *\emph{Tom bought anything} is not \citep{ladusaw1979polarity}.
Mandarin \zh{什么} `what' can also be read as `something', but only in licensing contexts. These contexts are not
the DE ones. Mandarin wh-indefinites are licensed by \emph{nonveridicality}, that is, by any context that does not
entail the event \citep{lin1998existential,xie2007,lin2015wh}. The two classes overlap under negation, questions and
conditionals, and they come apart in both directions:
\begin{itemize}\setlength\itemsep{0pt}
\item \zh{张伟昨天可能买了什么东西} `Zhang Wei perhaps bought something' is natural, but *\emph{Perhaps Tom bought
anything} is not: epistemic adverbs are nonveridical but not DE.
\item \emph{Only Tom bought anything} is natural, but \zh{只有张伟买了什么东西} is predicted to be degraded:
\emph{only} is Strawson-DE but entails that Zhang Wei bought something.
\end{itemize}
A model with one shared, English-shaped licensing detector would get the second contrast and miss the first.

We test this in a pre-registered study with twelve open LLMs from four families (0.5B--9B), and extend it in an
exploratory run to Qwen2.5-32B-Instruct. The items are 48 frames
in 21 contexts, with matched Mandarin and English versions. The measure is a log-probability difference-in-differences
that cancels frame plausibility and context length, normalised by negation. All \NMpass{} gated models show the
Mandarin side of the dissociation. Three LLM raters reading the same items show the same contrast. A second, exploratory protocol asks how the two
systems relate inside the model. Ablating a direction that carries English licensing removes English licensing in
\NEnenAbove{} of \NMech{} models, but Mandarin licensing beyond random in only \NEnzhAbove. Asked directly to choose
between sentences, most gated models also show the English side of the dissociation.

\paragraph{Contributions.}
\begin{itemize}\setlength\itemsep{0pt}
\item A controlled bilingual test of DE versus nonveridical licensing, released with code, items and all model
outputs. It covers contexts that existing Chinese benchmarks do not: existential \zh{什么}/\zh{谁} under
possibility, necessity and evidential adverbs.
\item A pre-registered finding that holds in all \NMpass{} gated models and both base models. Mandarin licensing
under nonveridical, non-DE adverbs is as strong as half to twice that of negation
($L_{\text{zh}}$ \LzhZAmin--\LzhZAmax), while English \emph{any} in the same contexts gets none
($L_{\text{en}}$ \LenZAmin--\LenZAmax). The effect survives two controls, two object families, an alternative
denominator and two adverb baselines (\NRobPos/\NRobAll{} cells).
\item An account of the weaker English side. \emph{Any} under \emph{only}/\emph{first}/\emph{regret} is licensed but
under-learned, and the Mandarin prediction in those contexts is itself context-dependent.
\item Two exploratory follow-ups, locked before their outputs. Directional ablation shows that the English
licensing direction stays in English in most models, while a Mandarin direction works in some families and not in
others. A prompted forced choice shows the full double dissociation in most gated models.
\end{itemize}

%% file: sections/02_setup.tex
\section{Design}
\label{sec:design}

\subsection{The contrast}

English \emph{any} is licensed by downward-entailing (DE) operators, including Strawson-DE ones such as \emph{only},
superlatives, and emotive factives \citep{ladusaw1979polarity,vonfintel1999npi}. Mandarin existential
\zh{什么}/\zh{谁} (`something'/`someone' on their indefinite reading) is licensed more broadly. Any
\emph{nonveridical} context licenses them, that is, any context that does not entail that the event happened
\citep{lin1998existential,xie2007,lin2015wh}. Nonveridicality includes DE but goes beyond it.
Two predictions follow, one on each side:
\begin{itemize}\setlength\itemsep{0pt}
\item \textbf{zh-advantage (ZA).} Epistemic possibility (\zh{可能} `perhaps'), necessity (\zh{肯定} `surely') and
evidentials (\zh{好像} `apparently') are nonveridical but not DE. They should license \zh{什么} but not \emph{any}:
\zh{张伟昨天可能买了什么东西} `Zhang Wei perhaps bought something' versus *\emph{Perhaps Tom bought anything}.
\item \textbf{en-advantage (EA).} \emph{Only}, \emph{the first to} and \emph{regret} are Strawson-DE, but they
entail that the event happened. They should license \emph{any} but not \zh{什么}, since
\zh{只有张伟买了} `only Zhang Wei bought' entails that he bought \citep[the non-entailment-of-existence
condition,][]{lin1998existential}.
\end{itemize}
A single English-shaped NPI detector predicts the English pattern in both languages. Two language-specific
licensing systems predict the double dissociation.

\subsection{Items}

Each item is a frame placed in one of 21 contexts (Table~\ref{tab:contexts} in the appendix). There are 48 frames:
24 with a thing object (\zh{买了什么东西} / \emph{bought anything}) and 24 with a person object
(\zh{见了谁} / \emph{met anyone}). Each frame has three objects:
\begin{itemize}\setlength\itemsep{0pt}
\item the target T (\zh{什么东西}, \zh{谁}; \emph{anything}, \emph{anyone});
\item a definite control C1 (\zh{那些东西}, \zh{那些人}; \emph{those things}, \emph{those people});
\item an indefinite control C2 (\zh{东西}, \zh{某个人}; \emph{something}, \emph{someone}).
\end{itemize}
This gives $48 \times 21 \times 3 \times 2 = 6{,}048$ sentences per model.

The contexts form seven classes:
\begin{itemize}\setlength\itemsep{0pt}
\item an episodic baseline (BASE);
\item four veridical adverbs (VERc: \emph{indeed}, \emph{predictably}, \emph{finally}, \emph{already});
\item two manner adverbs (MAN: \emph{secretly}, \emph{hastily});
\item the three shared licensers, negation (NEG), polar questions (QYN) and conditionals (COND);
\item the ZA classes POSS (4 adverbs), NEC (2) and EVID (2);
\item the three EA contexts.
\end{itemize}
The two adverb classes test whether a ZA effect merely reflects an extra preverbal adverb.

All contexts except QYN and COND, which have their own endings, close with the same continuation
(\zh{，然后就回家了} / \emph{and then went home}). The
continuation forces the declarative parse, because without it a bare \zh{张伟昨天买了什么东西} is a perfectly good
\emph{question}. Pilot items without it allowed an embedded-question parse, so it was fixed before the main study.

\subsection{Measure}

For frame $f$ and context $c$, let $\mathrm{LP}$ be the summed log-probability of the whole sentence. We score with
the model's BOS token, in bf16, and without a chat template. The licensing effect is a difference-in-differences:
\[
\begin{aligned}
\mathrm{DiD}(f,c) = {} & [\mathrm{LP}(\text{T}) - \mathrm{LP}(\text{C1})]_{f,c} \\
 - {} & [\mathrm{LP}(\text{T}) - \mathrm{LP}(\text{C1})]_{f,\text{BASE}}.
\end{aligned}
\]
The control C1 removes everything that the target and the control share, such as frame plausibility and context
length. The BASE term removes the target's baseline cost.

We call this relative effect \emph{licensing} throughout; it is graded and does not by itself separate grammatical
from ungrammatical sentences. We normalise by negation, $L(c) = \overline{\mathrm{DiD}}(c) / \overline{\mathrm{DiD}}(\text{NEG})$, so $L=1$ means
``as licensed as under negation''. The cross-language contrast is $\Delta(c) = L_{\text{zh}}(c) - L_{\text{en}}(c)$.

Contexts within a class are averaged frame by frame. Uncertainty comes from a frame bootstrap stratified by object
family (10{,}000 resamples, seed~0). Each resample recomputes the denominator.

\subsection{Pre-registered criteria}

The protocol, criteria and model list were committed before any main-study output existed. Only report-only
amendments were added later, and each one is dated in the released log.

\textbf{Measurement gate (M).} NEG and QYN must have $\mathrm{DiD}>0$ in both languages, with a CI that excludes~0
and $\geq$70\% of frames positive. Models that fail M are reported but not counted.

\textbf{ZA($k$)}, for $k \in$ \{POSS, NEC, EVID\}, requires all of the following:
\begin{itemize}\setlength\itemsep{0pt}
\item $\Delta(k)>0$, with a CI that excludes~0 and $\geq$60\% of frames positive;
\item $L_{\text{zh}}(k)>0$, with a CI that excludes~0;
\item $L_{\text{zh}}(k)$ above both $L_{\text{zh}}(\text{VERc})$ and $L_{\text{zh}}(\text{MAN})$, each with a CI that
excludes~0.
\end{itemize}

\textbf{EA($k$)}, for $k \in$ \{ONLY, FIRST, REGRET\}, is the mirror image: $\Delta(k)<0$ and
$L_{\text{en}}(k)>0$.

A model's zh side holds if ZA passes in $\geq$2 of the 3 classes; its en side holds if EA passes in $\geq$2 of 3.

\textbf{H1} requires $\geq$75\% of the gated models to show the zh side, including at least one model that is
neither Qwen nor Llama. \textbf{H2} requires $\geq$50\% to show the en side.

\subsection{Models}

We test twelve open models from four families. All are scored with Transformers~5 on single GPUs:
\begin{itemize}\setlength\itemsep{0pt}
\item Qwen2.5: 0.5B, 1.5B, 3B and 7B Instruct, plus 7B base;
\item Llama~3.x: 3.2-1B and 3.2-3B Instruct, 3.1-8B Instruct, plus 3.1-8B base;
\item Qwen3-8B;
\item two Chinese-centric models from other developers, GLM-4-9B-Chat and Yi-1.5-9B-Chat.
\end{itemize}

%% file: sections/03_results.tex
\section{Behavioural results}
\label{sec:results}

\input{tables/main}

\begin{figure}[t]
\centering
\includegraphics[width=\columnwidth]{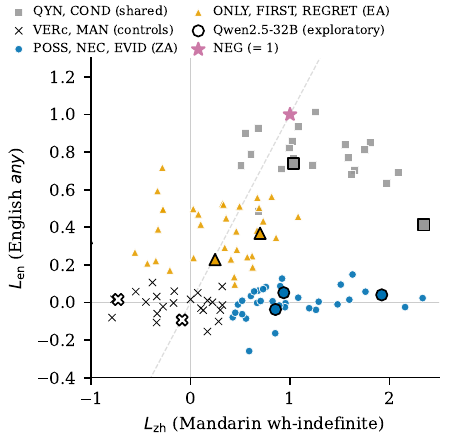}
\caption{Licensing strength in Mandarin ($L_{\text{zh}}$) against English ($L_{\text{en}}$), one point per gated
model and context class. Negation is 1 on both axes by definition; the dashed line is $L_{\text{zh}} =
L_{\text{en}}$. Nonveridical, non-DE adverbs (ZA, blue) license Mandarin wh-indefinites but not English \emph{any}.
DE, existence-entailing contexts (EA, orange) license \emph{any} weakly, and Mandarin \zh{什么} in some models.
Black outlines mark the exploratory Qwen2.5-32B-Instruct, which falls in the same regions.}
\label{fig:main}
\end{figure}

\paragraph{Gate.} \NMpass{} of the \NModels{} models pass the measurement gate. Qwen3-8B fails: its Mandarin
negation DiD is only \QthreeNegDiD{} (95\% CI [\QthreeNegLo, \QthreeNegHi]), with \QthreeNegFrac\% of frames
positive, while its QYN and COND cells behave normally. Because $L$ divides by the negation cell, its $L_{\text{zh}}$
values are not interpretable. Following the protocol, we report it (Table~\ref{tab:main}) but do not count it.

\paragraph{The Mandarin side holds everywhere (H1).} All \NzhSide{} gated models show the zh side, and every one
passes all three ZA classes (Figure~\ref{fig:main}, blue). This includes GLM-4 and Yi-1.5, the two models outside the Qwen and Llama families. The
effect is large:
\begin{itemize}\setlength\itemsep{0pt}
\item $L_{\text{zh}}$ ranges from \LzhZAmin{} to \LzhZAmax{}, so possibility, necessity and evidential adverbs license
\zh{什么}/\zh{谁} at roughly half to twice the strength of negation;
\item $L_{\text{en}}$ in the same contexts ranges from \LenZAmin{} to \LenZAmax{}, so \emph{perhaps}, \emph{surely}
and \emph{apparently} do nothing for \emph{any}.
\end{itemize}
The effect is not an adverb artefact: all \NAdvPos/\NAdvAll{} comparisons against the veridical and manner adverb
classes are positive with CIs excluding~0. The sign of $\Delta$ is also stable across all
\NRobPos/\NRobAll{} model $\times$ class $\times$ variant cells. The variants are the C1 and C2 controls, a
three-licenser denominator, and the thing and person families on their own.

\paragraph{The English side is weaker and model-dependent (H2 not met).} Only \NenSide{} of the \NMpass{} models
show the en side, against the pre-registered 50\%. The failure is not that \emph{any} goes unlicensed. $L_{\text{en}}$
is reliably above~0 under \emph{only} (\NOnlyLenPos/\NMpass{} models), \emph{the first to}
(\NFirstLenPos/\NMpass) and \emph{regret} (\NRegretLenPos/\NMpass). Its size, however, is only \LenEAmin--\LenEAmax{}
of negation. Two things make $\Delta$ fail:
\begin{itemize}\setlength\itemsep{0pt}
\item English licensing is weak in these contexts (gated-model means: ONLY \MeanLOnlyEn, FIRST \MeanLFirstEn,
REGRET \MeanLRegretEn);
\item many models also give Mandarin \zh{什么} some credit there (ONLY \MeanLOnlyZh, FIRST \MeanLFirstZh,
REGRET \MeanLRegretZh).
\end{itemize}
$\Delta<0$ holds in only \NOnlyDeltaNeg, \NFirstDeltaNeg{} and \NRegretDeltaNeg{} models respectively. Following the
protocol, the claim narrows to its pre-registered fallback. Mandarin nonveridical licensing is robust across
families and scales; the English DE side varies by model.

\paragraph{Scale and post-training.} The zh-side result has no scale threshold: it holds from 0.5B to 9B. It also
holds for both base models (Qwen2.5-7B, Llama-3.1-8B), so it is not a product of instruction tuning. An exploratory
extension outside the pre-registered list runs Qwen2.5-32B-Instruct through the unchanged pipeline, with its reading
fixed before any output. The model passes the gate, and the Mandarin side extends to it: all \BigNZA{} ZA classes are
licensed, at \BigLzhZAmin--\BigLzhZAmax{} times negation, while English \emph{any} gets at most \BigLenZAmax{} there.
The English side does not improve with scale. Only \BigNEA{} of the three EA classes holds, and mean English licensing
under EA (\BigLenEA) lies within the range of the 0.5B--7B Qwen2.5 models (\SmallLenEAmin--\SmallLenEAmax).

%% file: tables/main.tex
\begin{table*}[t]
\centering
\small
\setlength{\tabcolsep}{4pt}
\resizebox{\textwidth}{!}{\begin{tabular}{l rrr rrr c rrr rrr c}
\toprule
 & \multicolumn{7}{c}{Nonveridical, not DE (prediction: zh $+$, en $-$)} & \multicolumn{7}{c}{DE, existence-entailing (prediction: zh $-$, en $+$)}\\
\cmidrule(lr){2-8}\cmidrule(lr){9-15}
 & \multicolumn{3}{c}{$L_{\text{zh}}$} & \multicolumn{3}{c}{$L_{\text{en}}$} & & \multicolumn{3}{c}{$L_{\text{zh}}$} & \multicolumn{3}{c}{$L_{\text{en}}$} & \\
\cmidrule(lr){2-4}\cmidrule(lr){5-7}\cmidrule(lr){9-11}\cmidrule(lr){12-14}
Model & POSS & NEC & EVID & POSS & NEC & EVID & zh & ONLY & FIRST & REGR & ONLY & FIRST & REGR & en \\
\midrule
Qwen2.5-0.5B-Inst & 0.92 & 0.73 & 1.63 & 0.13 & 0.07 & 0.15 & \checkmark & 0.57 & 0.49 & 0.47 & 0.24 & 0.51 & 0.28 &  \\
Qwen2.5-1.5B-Inst & 0.68 & 0.61 & 1.76 & 0.06 & 0.03 & 0.06 & \checkmark & 0.03 & $-$0.28 & 0.86 & 0.50 & 0.50 & 0.34 &  \\
Qwen2.5-3B-Inst & 0.56 & 0.42 & 0.95 & $-$0.09 & $-$0.08 & $-$0.00 & \checkmark & 0.44 & $-$0.33 & 0.46 & 0.10 & 0.59 & 0.38 &  \\
Qwen2.5-7B-Inst & 0.76 & 0.48 & 1.51 & 0.08 & $-$0.01 & 0.10 & \checkmark & 0.62 & 0.36 & $-$0.43 & 0.19 & 0.45 & 0.21 &  \\
Qwen2.5-7B & 0.70 & 0.45 & 1.26 & 0.01 & $-$0.05 & $-$0.04 & \checkmark & 0.67 & 0.69 & $-$0.21 & 0.38 & 0.50 & 0.17 &  \\
\addlinespace[2pt]
Llama-3.2-1B-Inst & 0.64 & 0.52 & 0.90 & 0.06 & 0.01 & 0.09 & \checkmark & $-$0.33 & 0.82 & 0.41 & 0.39 & 0.56 & 0.23 &  \\
Llama-3.2-3B-Inst & 0.52 & 0.67 & 1.29 & $-$0.06 & $-$0.00 & 0.00 & \checkmark & 0.08 & 0.67 & 0.44 & 0.47 & 0.56 & 0.14 &  \\
Llama-3.1-8B-Inst & 0.82 & 1.08 & 1.48 & 0.01 & 0.03 & $-$0.01 & \checkmark & 0.10 & 0.32 & 0.34 & 0.42 & 0.53 & 0.29 & \checkmark \\
Llama-3.1-8B & 0.95 & 0.89 & 1.60 & $-$0.03 & $-$0.02 & 0.01 & \checkmark & $-$0.28 & 0.33 & 0.03 & 0.50 & 0.52 & 0.24 & \checkmark \\
\addlinespace[2pt]
GLM-4-9B-Chat & 1.19 & 0.88 & 2.33 & $-$0.03 & $-$0.02 & 0.02 & \checkmark & 0.73 & 1.08 & $-$0.56 & 0.43 & 0.46 & 0.27 &  \\
Yi-1.5-9B-Chat & 0.85 & 0.59 & 2.16 & $-$0.16 & $-$0.26 & $-$0.02 & \checkmark & 0.62 & $-$0.28 & $-$0.35 & 0.25 & 0.72 & 0.22 & \checkmark \\
Qwen3-8B$^\dagger$ & 0.28 & $-$3.34 & 1.93 & 0.11 & 0.12 & 0.13 &  & $-$6.94 & $-$3.43 & $-$3.51 & 0.10 & 0.58 & 0.15 &  \\
\bottomrule
\end{tabular}}
\caption{Licensing strength $L$ (DiD relative to negation; $L=1$ means as licensed as under negation, $L=0$ no licensing beyond the episodic baseline). zh/en: the model passes the pre-registered zh-side (ZA in $\geq$2 of 3 classes) or en-side (EA in $\geq$2 of 3) criterion. $^\dagger$Fails the measurement gate (its Mandarin negation cell collapses), so it is reported but not counted.}
\label{tab:main}
\end{table*}

%% file: sections/04_ratings.tex
\section{Do the items mean what the theory says?}
\label{sec:ratings}

The predictions come from the theoretical literature. This section asks how the target sentences actually read,
using two report-only checks that were added to the protocol before all models were in. Neither check enters the
criteria.

\paragraph{Native-speaker screening.} A native Mandarin speaker (one of the authors) screened all 48 frames and rated
168 target sentences on a 1--5 scale. No frame was flagged as unnatural, so the planned sensitivity analysis that
drops flagged frames is identical to the main analysis.

The screener rated almost every sentence as acceptable, including the EA contexts. We then showed the screener the 11 sentences that all three LLM raters scored $\leq 2$. They
judged 8 of them acceptable once a discourse context was supplied, for example \zh{昨天只有张伟见了谁} and
\zh{高翔是昨天第一个推谁的人}. In each such context, \zh{谁} picks out a particular person, such as the person one
had been hoping to see. The other 3 sentences stayed degraded: \zh{已经见了谁} and two \zh{第一个见/遇到谁的人}
sentences. We return to this
context sensitivity in \S\ref{sec:limits}.

\paragraph{LLM ratings.} Three LLM raters from different developers rated all 2{,}016 target sentences and all
2{,}016 C1 controls (temperature~0, shuffled batches of 12). We treat these as LLM ratings, not human judgements.
Inter-rater Spearman correlations per sentence are \AgreeZhMin--\AgreeZhMax{} for Mandarin and
\AgreeEnMin--\AgreeEnMax{} for English.

Averaged over raters, the ratings reproduce the theoretical contrast:
\begin{itemize}\setlength\itemsep{0pt}
\item Mandarin targets under POSS/NEC/EVID score close to the shared licensers, while English \emph{any} there scores
near the floor (\RateBaseEn{} on 1--5 at BASE);
\item English \emph{any} under ONLY/FIRST/REGRET scores \RateOnlyEn/\RateFirstEn/\RateRegretEn, while the Mandarin
targets score \RateOnlyZh/\RateFirstZh/\RateRegretZh.
\end{itemize}
Across the 20 non-baseline contexts, the mean rating tracks the gated models' mean $L$, with Spearman
$\rho=\RateRhoZh$ for Mandarin and $\RateRhoEn$ for English. A rating DiD built like the model DiD gives
$\rho=\RateDidRhoZh$ and $\RateDidRhoEn$.

The ratings also locate the English shortfall. The raters treat \emph{any} under \emph{only}, \emph{the first to}
and \emph{regret} as nearly as good as under negation. The models license it at only a fraction of negation. The
English side therefore fails because the models under-license Strawson-DE \emph{any}, not because the items are bad.

%% file: sections/05_mech.tex
\section{Is the licensing signal shared across languages?}
\label{sec:mech}

The behavioural dissociation leaves open how the two systems relate inside the model. An English-shaped detector
could still exist and be overridden for Mandarin, or the two languages could rely on separate signals. We test this
with directional interventions and a prompted forced choice. This second protocol was locked after the behavioural
verdict and before any of its outputs.

\paragraph{Directions.} At every layer $\ell$ we take the residual stream at the last token before the object,
$h_\ell$. For language $g$, the licensing direction is
\[
d_g(\ell) = \overline{h_\ell}[\text{LIC}] - \overline{h_\ell}[\text{UNL}],
\]
where LIC is NEG, QYN and COND and UNL is BASE, VERc and MAN.
This is the difference of means between the shared licensers and the unlicensed contexts, paired over the same 48
frames. The held-out classes, POSS/NEC/EVID and ONLY/FIRST/REGRET, never enter a direction. Three kinds of control
direction are also built:
\begin{itemize}\setlength\itemsep{0pt}
\item a lexical check $d_g^{\neg\text{NEG}}$, built from QYN and COND only and tested on the negation cell it never
saw;
\item an adverb control $d_g^{\text{VER}}$ = VERc $-$ BASE;
\item 20 random Gaussian directions per layer.
\end{itemize}

\paragraph{Interventions.} \emph{Ablation} projects each layer's own unit direction $u_g(\ell)$ out of that layer's
block output, $h \leftarrow h - (h\cdot u)u$, at every position and every layer. A sensitivity variant ablates only
the first half of the layers. The effect on class $k$ is the removal share
$R(k) = 1 - \overline{\mathrm{DiD}}_{\text{abl}}(k) / \overline{\mathrm{DiD}}(k)$. \emph{Addition} adds
$\alpha u_g$ at the middle layer to BASE sentences, with $\alpha$ set to the norm of the target language's
$d_g$ at that layer.

The pre-registered test ablates $u_{\text{en}}$ and asks what happens to Mandarin. A shared English-shaped signal
predicts that $u_{\text{en}}$ removes Mandarin negation licensing and, to a similar degree, Mandarin ZA licensing.
The categories are:
\begin{itemize}\setlength\itemsep{0pt}
\item \emph{shared}: $R_{\text{NEG}}$ and $R_{\text{ZA}}$ are both above the random 95th percentile, and $R_{\text{ZA}}$
is not more than 0.25 below $R_{\text{NEG}}$ (bootstrap CI);
\item \emph{separate}: $R_{\text{NEG}}$ is above the random 95th percentile, but $R_{\text{ZA}}$ is not, or $R_{\text{ZA}}$
is more than 0.25 below $R_{\text{NEG}}$ (CI);
\item \emph{neither}: $R_{\text{NEG}}$ is not above the random 95th percentile;
\item \emph{partial}: anything else.
\end{itemize}
A category describes the model set if it holds in a majority of the assessable models.

\paragraph{Measure, and an amendment.} The protocol scored only the object word given the prefix. The first model's
measurement check showed that this misses most of the Mandarin effect: across models it tracks the behavioural DiD
only at $\rho$ = \ChkOZhMin--\ChkOZhMax{} for Mandarin. The reason is linguistic. \zh{张伟昨天买了什么东西} is a
fine question, so \zh{什么} is not dispreferred at the object; the cost arrives with the declarative continuation.
(For QYN the Mandarin prefix equals BASE, so the object-only DiD is zero by construction.) After inspecting that
first model we amended the protocol to also score object plus continuation, and reran all models. This measure
scores the same tokens as the behavioural DiD, so its agreement with it ($\rho \geq \ChkFZhMin$) confirms the
alignment rather than replicating the result. Directions, layers, random baselines and categories are
unchanged. We report both measures below; the continuation measure is the post-hoc one.

\begin{figure*}[t]
\centering
\includegraphics[width=\textwidth]{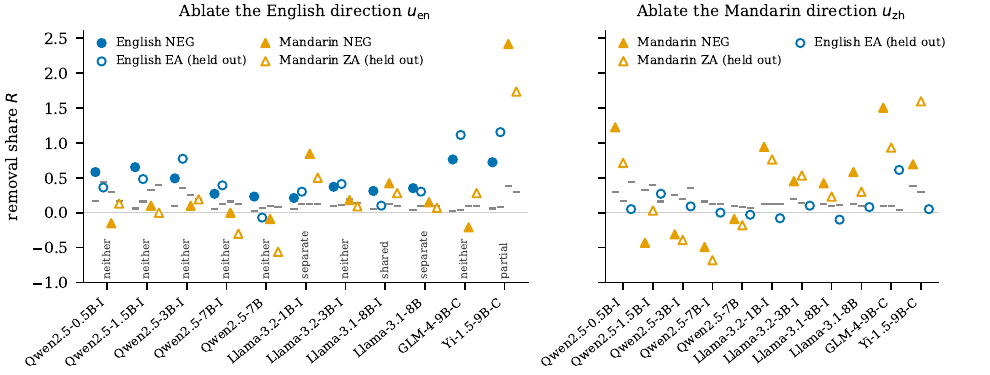}
\caption{Directional ablation per model, on the continuation measure. Each point is a removal share $R$; the grey
dash next to it is the 95th percentile of the same share over 20 random directions. Filled markers are the
negation cell, which the direction was built with; hollow markers are held-out classes (ZA: POSS/NEC/EVID; EA:
ONLY/FIRST/REGRET). The word under each model in the left panel is its pre-registered category. Values are in
Table~\ref{tab:mech}; the object-only measure is in Table~\ref{tab:mechlocked}.}
\label{fig:mech}
\end{figure*}

\paragraph{Results: the English direction stays in English.} Figure~\ref{fig:mech} (left) shows the test.
Ablating $u_{\text{en}}$ removes English negation licensing beyond random in \NEnenAbove{} of \NMech{} models
($R$ = \RenenMin--\RenenMax), and it also removes the held-out EA licensing in \NEnenEAAbove. The direction
therefore captures English licensing, including classes it was not built from. The same direction removes
Mandarin negation licensing beyond random in only \NEnzhAbove{} models, and Mandarin ZA licensing in
\NEnzhZAAbove. Under the pre-registered categories the model set reads \emph{neither}: \NCatFNeither{} of
\NCatFAssess{} models, against \NCatFShared{} shared, \NCatFSeparate{} separate and \NCatFPartial{} partial. Ablating
only the first half of the layers gives the same reading (\NCatEFNeither{} neither, \NCatEFShared{} shared). On
the pre-specified object-only measure the Mandarin NEG or ZA effect is not positive in \NCatONotassessable{} models,
which therefore cannot be classified. No category reaches a majority of the other \NCatOAssess{} when all layers are
ablated; with the first half ablated, \NCatEONeither{} of them read neither (Table~\ref{tab:mechlocked}).

\paragraph{The Mandarin direction is model-dependent.} The null for $u_{\text{en}}$ means something only where a
Mandarin direction works at all (Figure~\ref{fig:mech}, right). Ablating $u_{\text{zh}}$ removes Mandarin negation
and held-out ZA licensing together in \NZhzhBoth{} models: all Llama, GLM and Yi models and Qwen2.5-0.5B. In
Qwen2.5-1.5B to 7B it removes neither, so in these models our difference-of-means direction has no detectable effect on
Mandarin licensing, and their \emph{neither} says nothing about sharing. Among the \NZhzhBoth{} models where $u_{\text{zh}}$
works, only Llama-3.1-8B-Inst is shared. In the others the English direction misses Mandarin negation (three models),
removes it without the ZA classes (two), or removes both by clearly different amounts (Yi-1.5). Our conclusion is therefore limited: the English licensing direction
reliably affects English licensing, but it does not reliably reach Mandarin licensing, and the Mandarin direction's
effect depends on the model. This does not exclude a shared representation that is non-linear or sits elsewhere.

\paragraph{Addition.} Adding a direction to BASE sentences at the middle layer raises English \emph{any} above
random for $u_{\text{en}}$ in \NAddEnen{} models, and also for $u_{\text{zh}}$ in \NAddZhen. Mandarin \zh{什么} rarely
rises above random ($u_{\text{zh}}$: \NAddZhzh{} model; $u_{\text{en}}$: \NAddEnzh). The addition test is thus uninformative for
Mandarin, and the English result suggests that both directions share a generic licenser component, such as
negation, that English \emph{any} responds to.

\paragraph{Prompted forced choice.} The second protocol also asks each model which of two sentences is more
natural, with both orders averaged. \NFcGate{} models pass the negation gate in both languages. In these models the
preferences show the full double dissociation. The Mandarin advantage under the ZA classes has a CI above zero in
\NFcPossPos, \NFcNecPos{} and \NFcEvidPos{} of the \NFcGate{} models (POSS, NEC, EVID). The English advantage under
\emph{only}, \emph{first} and \emph{regret} has a CI below zero in \NFcOnlyNeg, \NFcFirstNeg{} and \NFcRegretNeg.
The prompted $L$ values track the log-probability $L$ values ($\rho$ = \FcRhoMin--\FcRhoMax). These are prompted
preferences of the same models, not human judgements, but they show that the English side, weak in sentence
probabilities, is present when the models are asked directly. The exploratory 32B model shows the full double
dissociation in forced choice as well.

%% file: sections/06_related.tex
\section{Related work}
\label{sec:related}

\paragraph{Polarity items in language models.} Most of this work is about English.
\begin{itemize}\setlength\itemsep{0pt}
\item NPI licensing was among the first targeted syntactic evaluations
\citep{marvin2018targeted,jumelet2018npi,warstadt2019npi}.
\item BLiMP includes NPI paradigms \citep{warstadt2020blimp}.
\item \citet{jumelet2021monotonicity} show that LMs use monotonicity information to judge NPI licensing.
\item \citet{bylinina2022polarity} probe polarity sensitivity in neural LMs.
\item \citet{kumon2026shared} use activation patching on English constructions. Filler-gap dependencies share a
localised mechanism across constructions, but NPI licensing does not.
\end{itemize}
Chinese benchmarks cover \zh{任何} `any' under negation, conditionals, questions and non-factive verbs
\citep{liu2026zhoblimp}, as well as \zh{什么} under negation \citep{song2022sling}. The multilingual BLiMPPI covers
polarity items in 33 languages but excludes conditionals \citep{guven2026blimppi}. None of these tests existential
\zh{什么}/\zh{谁} in nonveridical contexts that are not DE. None contrasts DE with nonveridical licensing within one
design, or sets Mandarin against English.

\paragraph{Theory.} \citet{ladusaw1979polarity} tied NPIs to downward entailment, and \citet{vonfintel1999npi}
extended this to Strawson-DE operators such as \emph{only}. \citet{giannakidou1998polarity} argued that many polarity
items depend on nonveridicality instead. For Mandarin wh-indefinites, \citet{lin1998existential} proposed a
non-entailment-of-existence condition, and \citet{xie2007} and \citet{lin2015wh} recast it as nonveridical
licensing. \citet{chen2017epistemic} shows that \zh{什么} can also appear in some veridical episodic sentences, where
it carries a speaker-ignorance reading. This matters for our en-advantage cells (\S\ref{sec:discussion}).

\paragraph{Shared multilingual representations.} Latent representations in multilingual LMs are partly
English-anchored \citep{wendler2024llamas}. For negation specifically, \citet{nguyen2026not} find a shared
cross-lingual representation: a direction extracted in one language adds or removes negation in others. Our
intervention asks the next question. Is the \emph{licensing} that negation, questions and conditionals provide also
carried by a shared direction, and does such a direction cover the Mandarin-specific licensers? We use
difference-of-means directions with directional ablation and addition, following \citet{arditi2024refusal}. Our
forced-choice probe follows the caution of \citet{hu2023prompting} that prompted judgements and probabilities can
come apart.

%% file: sections/07_discussion.tex
\section{Discussion}
\label{sec:discussion}

\paragraph{What the models learned.} Every gated model, from 0.5B to 9B in four families and the exploratory 32B
model, licenses Mandarin
\zh{什么}/\zh{谁} under possibility, necessity and evidential adverbs. None licenses English \emph{any} there. In
these contexts both languages offer the same surface cue, a preverbal adverb, yet only Mandarin licenses the target.
A single English-calibrated NPI detector cannot produce this. In Mandarin, the models' scores follow the property that the
theory says matters there, nonveridicality, rather than the DE property that governs \emph{any}.
The ablations point the same way. A direction that carries English licensing, held-out classes included, does not
reach Mandarin licensing in most models (Section~\ref{sec:mech}).

\paragraph{Why the English side is weaker.} Two separate things happen in the en-advantage cells.
\begin{itemize}\setlength\itemsep{0pt}
\item \emph{Any} under \emph{only}, \emph{the first to} and \emph{regret} is licensed, but at a fraction of negation.
LLM raters, by contrast, accept these sentences almost as readily as negated ones. Strawson-DE licensing is the
part of the English system that these models under-learn, at least at the scales we test: in Qwen2.5, the 32B model
is no better at it than the 0.5B--7B models. This fits its rarity and the presupposition it requires
\citep{vonfintel1999npi}.
\item Several models give Mandarin \zh{什么} partial credit under \zh{只有} and \zh{第一个}. That is not obviously a
model error. Our native screener accepted such sentences once a context made the wh-phrase refer to a particular
person or thing. \citet{chen2017epistemic} documents a speaker-ignorance use of \zh{什么} in veridical sentences.
The non-entailment-of-existence prediction for these cells is therefore weaker than the nonveridical prediction for
the ZA cells, and the asymmetry in our results mirrors that.
\end{itemize}

\section*{Limitations}
\label{sec:limits}

\begin{itemize}\setlength\itemsep{1pt}
\item \textbf{Relative, not categorical, licensing.} $L$ measures how much a context raises the target over a
definite control, relative to negation. It does not separate grammatical from ungrammatical. The native-speaker
follow-up shows that several ``unlicensed'' Mandarin sentences improve in a supporting context, so the EA-side
predictions are graded.
\item \textbf{Human judgements.} The only human check is a screening by one native speaker, who is an author. The
three LLM raters reproduce the theoretical contrasts, but they are LLM ratings, not human data. A rating study with
naive native speakers of both languages remains to be done.
\item \textbf{Items.} The design uses 48 templated frames with one continuation, so the effects are measured in a
controlled but narrow construction. Naturalistic sentences, other wh-words and embedded contexts (for example,
non-factive verbs) are untested.
\item \textbf{Models.} The pre-registered models go up to 9B. One exploratory 32B model, from the Qwen2.5 family,
extends this; other families at that scale and larger models are untested. One model, Qwen3-8B, fails the measurement gate because its
Mandarin negation cell collapses. We report it rather than adjusting the gate.
\item \textbf{Mechanism.} The interventions use linear difference-of-means directions read at one position. A
null effect for a direction does not show that no shared representation exists, and in Qwen2.5-1.5B to 7B even the
Mandarin direction has no effect on Mandarin. The continuation measure was added after the first model's
measurement check; on the pre-specified object-only measure the cross-model reading is inconclusive.
\end{itemize}

%% file: sections/09_appendix.tex
\section{Items}
\label{app:items}

\input{tables/contexts}

\section{Ablation values}
\label{app:locked}

Table~\ref{tab:mech} gives the values plotted in Figure~\ref{fig:mech}. Table~\ref{tab:mechlocked} gives the same
ablation on the pre-specified object-only measure. Rerunning that measure with the amended code reproduces the first
run (Pearson $r \geq \VvRhoMin$ on the three models run twice).

\input{tables/mech}

\input{tables/mech_locked}

%% file: tables/contexts.tex
\begin{table*}[t]
\centering
\footnotesize
\setlength{\tabcolsep}{3pt}
\resizebox{\textwidth}{!}{\begin{tabular}{l c l l}
\toprule
Context & zh/en & Mandarin (T) & English (T) \\
\midrule
BASE & $-$/$-$ & 张伟昨天买了什么东西\ldots & Tom bought anything yesterday\ldots \\
VER1 & $-$/$-$ & 张伟昨天确实买了什么东西\ldots & Tom indeed bought anything yesterday\ldots \\
VER2 & $-$/$-$ & 张伟昨天果然买了什么东西\ldots & Tom predictably bought anything yesterday\ldots \\
VER3 & $-$/$-$ & 张伟昨天终于买了什么东西\ldots & Tom finally bought anything yesterday\ldots \\
VER4 & $-$/$-$ & 张伟昨天已经买了什么东西\ldots & Tom already bought anything yesterday\ldots \\
MAN1 & $-$/$-$ & 张伟昨天偷偷买了什么东西\ldots & Tom secretly bought anything yesterday\ldots \\
MAN2 & $-$/$-$ & 张伟昨天匆匆买了什么东西\ldots & Tom hastily bought anything yesterday\ldots \\
NEG & $+$/$+$ & 张伟昨天没买什么东西\ldots & Tom didn't buy anything yesterday\ldots \\
QYN & $+$/$+$ & 张伟昨天买了什么东西吗？ & Did Tom buy anything yesterday? \\
COND & $+$/$+$ & 如果张伟昨天买了什么东西，请马上告诉我。 & If Tom bought anything yesterday, please tell me right away. \\
POSS1 & $+$/$-$ & 张伟昨天可能买了什么东西\ldots & Perhaps Tom bought anything yesterday\ldots \\
POSS2 & $+$/$-$ & 张伟昨天也许买了什么东西\ldots & Tom might have bought anything yesterday\ldots \\
POSS3 & $+$/$-$ & 张伟昨天大概买了什么东西\ldots & Tom probably bought anything yesterday\ldots \\
POSS4 & $+$/$-$ & 张伟昨天或许买了什么东西\ldots & Maybe Tom bought anything yesterday\ldots \\
NEC1 & $+$/$-$ & 张伟昨天肯定买了什么东西\ldots & Tom surely bought anything yesterday\ldots \\
NEC2 & $+$/$-$ & 张伟昨天一定买了什么东西\ldots & Tom must have bought anything yesterday\ldots \\
EVID1 & $+$/$-$ & 张伟昨天好像买了什么东西\ldots & Tom apparently bought anything yesterday\ldots \\
EVID2 & $+$/$-$ & 张伟昨天似乎买了什么东西\ldots & Tom seemingly bought anything yesterday\ldots \\
ONLY & $-$/$+$ & 昨天只有张伟买了什么东西\ldots & Only Tom bought anything yesterday\ldots \\
FIRST & $-$/$+$ & 张伟是昨天第一个买什么东西的人\ldots & Tom was the first to buy anything yesterday\ldots \\
REGRET & $-$/$+$ & 张伟后悔昨天买了什么东西\ldots & Tom regretted having bought anything yesterday\ldots \\
\bottomrule
\end{tabular}}
\caption{The 21 contexts, shown for thing frame~0 with the target object; the second column gives the predicted licensing. \ldots{} stands for the shared continuation \zh{，然后就回家了。} / \emph{, and then went home.} The controls replace the object with \zh{那些东西}/\emph{those things} (C1) or \zh{东西}/\emph{something} (C2). Person frames use \zh{谁}, \zh{那些人}, \zh{某个人} and \emph{anyone}, \emph{those people}, \emph{someone}.}
\label{tab:contexts}
\end{table*}

%% file: tables/mech.tex
\begin{table*}[t]
\centering
\small
\setlength{\tabcolsep}{4pt}
\resizebox{\textwidth}{!}{\begin{tabular}{l rr rr rr rr rr r l}
\toprule
 & \multicolumn{6}{c}{Mandarin licensing removed} & \multicolumn{5}{c}{English licensing removed} & \\
\cmidrule(lr){2-7}\cmidrule(lr){8-12}
 & \multicolumn{2}{c}{by $u_{\text{en}}$} & \multicolumn{2}{c}{random r95} & \multicolumn{2}{c}{by $u_{\text{zh}}$} & \multicolumn{2}{c}{by $u_{\text{en}}$} & \multicolumn{2}{c}{random r95} & by $u_{\text{zh}}$ & \\
\cmidrule(lr){2-3}\cmidrule(lr){4-5}\cmidrule(lr){6-7}\cmidrule(lr){8-9}\cmidrule(lr){10-11}\cmidrule(lr){12-12}
Model & NEG & ZA & NEG & ZA & NEG & ZA & NEG & EA & NEG & EA & EA & category \\
\midrule
Qwen2.5-0.5B-Inst & $-$0.15 & 0.13 & 0.29 & 0.17 & 1.22 & 0.71 & 0.58 & 0.36 & 0.17 & 0.44 & 0.05 & neither \\
Qwen2.5-1.5B-Inst & 0.10 & $-$0.00 & 0.32 & 0.39 & $-$0.43 & 0.03 & 0.65 & 0.48 & 0.06 & 0.16 & 0.27 & neither \\
Qwen2.5-3B-Inst & 0.10 & 0.19 & 0.25 & 0.19 & $-$0.31 & $-$0.39 & 0.49 & 0.77 & 0.10 & 0.35 & 0.09 & neither \\
Qwen2.5-7B-Inst & $-$0.00 & $-$0.30 & 0.16 & 0.12 & $-$0.49 & $-$0.68 & 0.27 & 0.39 & 0.05 & 0.13 & $-$0.00 & neither \\
Qwen2.5-7B & $-$0.09 & $-$0.56 & 0.09 & 0.08 & $-$0.09 & $-$0.18 & 0.23 & $-$0.07 & 0.03 & 0.07 & $-$0.03 & neither \\
Llama-3.2-1B-Inst & 0.84 & 0.50 & 0.13 & 0.13 & 0.94 & 0.76 & 0.21 & 0.30 & 0.05 & 0.12 & $-$0.08 & separate \\
Llama-3.2-3B-Inst & 0.18 & 0.09 & 0.19 & 0.14 & 0.45 & 0.53 & 0.37 & 0.41 & 0.09 & 0.11 & 0.10 & neither \\
Llama-3.1-8B-Inst & 0.42 & 0.28 & 0.12 & 0.09 & 0.42 & 0.23 & 0.31 & 0.10 & 0.05 & 0.11 & $-$0.10 & shared \\
Llama-3.1-8B & 0.15 & 0.07 & 0.13 & 0.09 & 0.58 & 0.30 & 0.35 & 0.30 & 0.04 & 0.10 & 0.08 & separate \\
GLM-4-9B-Chat & $-$0.21 & 0.28 & 0.10 & 0.10 & 1.50 & 0.93 & 0.76 & 1.11 & 0.02 & 0.04 & 0.61 & neither \\
Yi-1.5-9B-Chat & 2.41 & 1.73 & 0.38 & 0.30 & 0.69 & 1.59 & 0.72 & 1.15 & 0.06 & 0.08 & 0.05 & partial \\
\bottomrule
\end{tabular}}
\caption{Directional ablation (every layer projects out its own direction). Entries are removal shares $R = 1 - \mathrm{DiD}_{\text{abl}}/\mathrm{DiD}$ on the sentence-continuation measure; r95 is the 95th percentile over 20 random directions. ZA averages POSS/NEC/EVID; EA averages ONLY/FIRST/REGRET. The directions are built only from NEG/QYN/COND versus unlicensed contexts, so ZA and EA are held out.}
\label{tab:mech}
\end{table*}

%% file: tables/mech_locked.tex
\begin{table*}[t]
\centering
\small
\setlength{\tabcolsep}{4pt}
\resizebox{\textwidth}{!}{\begin{tabular}{l rr rr rr rr rr r l}
\toprule
 & \multicolumn{6}{c}{Mandarin licensing removed} & \multicolumn{5}{c}{English licensing removed} & \\
\cmidrule(lr){2-7}\cmidrule(lr){8-12}
 & \multicolumn{2}{c}{by $u_{\text{en}}$} & \multicolumn{2}{c}{random r95} & \multicolumn{2}{c}{by $u_{\text{zh}}$} & \multicolumn{2}{c}{by $u_{\text{en}}$} & \multicolumn{2}{c}{random r95} & by $u_{\text{zh}}$ & \\
\cmidrule(lr){2-3}\cmidrule(lr){4-5}\cmidrule(lr){6-7}\cmidrule(lr){8-9}\cmidrule(lr){10-11}\cmidrule(lr){12-12}
Model & NEG & ZA & NEG & ZA & NEG & ZA & NEG & EA & NEG & EA & EA & category \\
\midrule
Qwen2.5-0.5B-Inst & 2.82 & 1.79 & 1.48 & 0.49 & 4.93 & 1.82 & 1.31 & 2.95 & 0.16 & 1.26 & 1.04 & partial \\
Qwen2.5-1.5B-Inst & $-$0.42 & $-$0.43 & 1.60 & 0.75 & 3.88 & 1.24 & 1.01 & 1.71 & 0.12 & 0.31 & 0.54 & neither \\
Qwen2.5-3B-Inst & $-$0.37 & $-$0.24 & 0.22 & 0.56 & $-$0.36 & 0.15 & 0.83 & 2.06 & 0.10 & 0.32 & 0.30 & neither \\
Qwen2.5-7B-Inst & 0.75 & $-$0.39 & 0.67 & 0.18 & 1.39 & 0.09 & 0.71 & 1.54 & 0.07 & 0.51 & $-$0.27 & separate \\
Qwen2.5-7B & 1.31 & 0.03 & 0.28 & 0.14 & 0.55 & 0.35 & 0.76 & 1.20 & 0.05 & 0.22 & $-$0.22 & separate \\
Llama-3.2-1B-Inst & 3.78 & 1.30 & 0.61 & 0.32 & 3.18 & 1.22 & 0.29 & 2.25 & 0.10 & 0.40 & 0.11 & partial \\
Llama-3.2-3B-Inst & -- & 1.13 & -- & 0.32 & -- & 1.38 & 0.72 & 1.87 & 0.09 & 0.28 & 0.70 & n/a \\
Llama-3.1-8B-Inst & -- & 0.20 & -- & 0.21 & -- & 1.46 & 0.83 & 1.42 & 0.07 & 0.20 & $-$0.05 & n/a \\
Llama-3.1-8B & 0.45 & 0.17 & 0.23 & 0.10 & 1.86 & 0.86 & 0.88 & 1.91 & 0.05 & 0.18 & 0.83 & partial \\
GLM-4-9B-Chat & -- & 0.90 & -- & 0.30 & -- & 2.46 & 1.05 & 2.14 & 0.04 & 0.13 & 1.72 & n/a \\
Yi-1.5-9B-Chat & -- & -- & -- & -- & -- & -- & 0.54 & 1.96 & 0.12 & 0.22 & 0.25 & n/a \\
\bottomrule
\end{tabular}}
\caption{The same ablation on the pre-specified object-only measure. It misses most of the Mandarin effect (Section~\ref{sec:mech}); -- marks classes whose unablated DiD is $\leq 0$ (not assessable).}
\label{tab:mechlocked}
\end{table*}